\documentclass[letterpaper,twocolumn,10pt]{article}
\usepackage{usenix,epsfig,endnotes}

\usepackage{lastpage,framed,url,subcaption,graphicx,cleveref}
\usepackage{amsmath,xcolor,soul,algpseudocode,algorithm,tabularx,multirow}
\usepackage{amssymb,cite,balance}
\usepackage{xspace} 

\PassOptionsToPackage{hyphens}{url}

\expandafter\def\expandafter\UrlBreaks\expandafter{\UrlBreaks
  \do\a\do\b\do\c\do\d\do\e\do\f\do\g\do\h\do\i\do\j%
  \do\k\do\l\do\m\do\n\do\o\do\p\do\q\do\r\do\s\do\t%
  \do\u\do\v\do\w\do\x\do\y\do\z\do\A\do\B\do\C\do\D%
  \do\E\do\F\do\G\do\H\do\I\do\J\do\K\do\L\do\M\do\N%
  \do\O\do\P\do\Q\do\R\do\S\do\T\do\U\do\V\do\W\do\X%
  \do\Y\do\Z}

\newcount\Comments  \Comments=1 
\newcommand{\kibitz}[2]{\ifnum\Comments=1\textcolor{#1}{#2}\fi}

\providecommand{\vs}{vs.\ }
\providecommand{\ie}{\emph{i.e.,}\xspace}
\providecommand{\eg}{\emph{e.g.,}\xspace}

\usepackage{xpunctuate}

\providecommand{\myparab}[1]{\smallskip\noindent\textbf{#1} }

\usepackage{booktabs}
\newenvironment{tab}[1]
{
\let\oldarraystretch=\arraystretch
\renewcommand{\arraystretch}{1.1} 
\begin{tabular}{@{}#1@{}}
\toprule
}
{\bottomrule
\end{tabular}
\renewcommand{\arraystretch}{\oldarraystretch}
}

\newcommand{\squishenum}{   \begin{enumerate}{}    { \setlength{\itemsep}{0pt}      \setlength{\parsep}{0pt}      \setlength{\topsep}{3pt}       \setlength{\partopsep}{0pt}      \setlength{\leftmargin}{1.5em} \setlength{\labelwidth}{1em}      \setlength{\labelsep}{0.5em} } }
\newcommand{\squishlist}{   \begin{list}{$\bullet$}    { \setlength{\itemsep}{0pt}      \setlength{\parsep}{3pt}      \setlength{\topsep}{3pt}       \setlength{\partopsep}{0pt}      \setlength{\leftmargin}{1.5em} \setlength{\labelwidth}{1em}      \setlength{\labelsep}{0.5em} } }
\newcommand{\squishlisttwo}{   \begin{list}{$\bullet$}    { \setlength{\itemsep}{0pt}    \setlength{\parsep}{0pt}      \setlength{\topsep}{0pt}     \setlength{\partopsep}{0pt}      \setlength{\leftmargin}{2em} \setlength{\labelwidth}{1.5em}      \setlength{\labelsep}{0.5em} } }
\newcommand{\squishend}{    \end{list}  }
\newcommand{\squishenumend}{	\end{enumerate}	}

\begin{document}
\date{}

\title{Measuring Offensive Speech in Online Political Discourse}
\author{
Rishab Nithyanand$^1$, Brian Schaffner$^2$, Phillipa Gill$^1$
\\
\\
$^1$\{rishab, phillipa\}@cs.umass.edu, $^2$schaffne@polsci.umass.edu\\
University of Massachusetts, Amherst
}

\pagenumbering{gobble}
\maketitle

\begin{abstract}
The Internet and online forums such as Reddit have become an increasingly
popular medium for citizens to engage in political conversations. However, the
online disinhibition effect resulting from the ability to use pseudonymous
identities may manifest in the form of offensive speech, consequently making
political discussions more aggressive and polarizing than they already are. Such
environments may result in harassment and self-censorship from its targets. In
this paper, we present preliminary results from a large-scale temporal
measurement aimed at quantifying offensiveness in online political discussions.

To enable our measurements, we develop and evaluate an offensive speech
classifier. We then use this classifier to quantify and compare offensiveness in
the political and general contexts. We perform our study using a database of
over 168M Reddit comments made by over 7M pseudonyms  between January 2015 and
January  2017 -- a period covering several divisive political events including
the 2016 US presidential elections.
\end{abstract}

\section{Introduction}\label{sec:introduction}

The apparent rise in political incivility has attracted substantial attention
from scholars in recent years. These studies have largely focused on the extent
to which politicians and elected officials are increasingly employing rhetoric
that appears to violate norms of civility \cite{herbst2010rude, mutz2015your}.
For the purposes of our work, we use the incidence of offensive rhetoric as a
stand in for incivility.
The 2016 US presidential election was an especially noteworthy case in this
regard, particularly in terms of Donald Trump's campaign which frequently
violated norms of civility both in how he spoke about broad groups in the public
(such as Muslims, Mexicans, and African Americans) and the attacks he leveled at
his opponents \cite{gross2016twitter}. The consequences of incivility are
thought to be crucial to the functioning of democracy since ``public civility
and interpersonal politeness sustain social harmony and allow people who
disagree with one another to maintain ongoing relationships"
\cite{strachan2012political}.

While political incivility appears to be on the rise among elites, it is less
clear whether this is true among the mass public as well. Is political discourse
particularly lacking in civility compared to discourse more generally? Does the
incivility of mass political discourse respond to the dynamics of political
campaigns? Addressing these questions has been difficult for political
scientists because traditional tools for studying mass behavior, such as public
opinion surveys, are ill-equipped to measure how citizens discuss politics with
one another. Survey data does reveal that the public tends to perceive politics
as becoming increasingly less civil during the course of a political campaign
\cite{wolf2012incivility}. Yet, it is not clear whether these perceptions match
the reality, particularly in terms of the types of discussions that citizens
have with each other. 

An additional question is how incivility is received by others. On one hand,
violations of norms regarding offensive discourse may be policed by members of a
community, rendering such speech ineffectual. On the other hand, offensive
speech may be effective as a means for drawing attention to a particular
argument. Indeed, there is evidence that increasing incivility in political
speech results in higher levels of attention from the public
\cite{mutz2015your}. During the 2016 campaign, the use of swearing in comments
posted on Donald Trump's YouTube channel tended to result in additional
responses that mimicked such swearing \cite{kwon2017aggression}. Thus, offensive
speech in online fora may attract more attention from the community and lead to
the spread of even more offensive speech in subsequent posts. 

To address these questions regarding political incivility, we examine the use of
offensive speech in political discussions housed on Reddit. Scholars tend to
define uncivil discourse as ``communication that violates the norms of
politeness" \cite{mutz2015your} a definition that clearly includes offensive
remarks. Reddit fora represent a ``most likely" case for the study of offensive
political speech due its strong free speech culture \cite{reddit-freespeech} and
the ability of participants to use pseudonymous identities. That is, if
political incivility in the public did increase during the 2016 campaign, this
should be especially evident on fora such as Reddit. Tracking Reddit discussions
throughout all of 2015 and 2016, we find that online  political discussions
became increasingly more offensive as the general election campaign intensified.
By comparison, discussions on non-political subreddits did not become
increasingly offensive during this period. Additionally, we find that the
presence of offensive comments did not subside even three months after the
election.


\section{Datasets}\label{sec:datasets}
Our study makes use of multiple datasets in order to identify and characterize
trends in offensive speech.

\myparab{The CrowdFlower hate speech dataset.} The CrowdFlower hate speech
dataset \cite{CrowdFlower-Hate} contains 14.5K tweets, each receiving labels
from at least three contributors. Contributors were allowed to classify each
tweet into one of three classes: {Not Offensive} (NO), {Offensive but not
hateful} (O), and {Offensive and hateful} (OH). Of the 14.5K tweets, only
37.6\% had a decisive class -- \ie the same class was assigned by all
contributors. For indecisive cases, the majority class was selected and a
class confidence score (fraction of contributors that selected the majority
class) was made available. Using this approach, 50.4\%, 33.1\%, and 16.5\% of
the tweets were categorized as NO, O, and OH, respectively. Since our goal is to
identify any offensive speech (not just hate speech), we consolidate assigned
classes into Offensive and Not Offensive by relabeling OH tweets as Offensive.
We use this modified dataset to train, validate, and test our offensive speech
classifier. To the best of our knowledge, this is the only dataset that provides
\emph{offensive} and \emph{not offensive} annotations to a large dataset.
 
\myparab{Offensive word lists.} We also use two offensive word lists as
auxiliary input to our classifier: (1) The Hatebase hate speech vocabulary
\cite{Hatebase-API} consisting of 1122 hateful words and (2) 422 offensive
words banned from Google's What Do You Love project \cite{Google-WDYL-data}. 

\begin{figure}[htb]
\centering
\includegraphics[trim=0cm 0cm 0cm 0cm, clip=true, width=.49\textwidth, angle=0]
{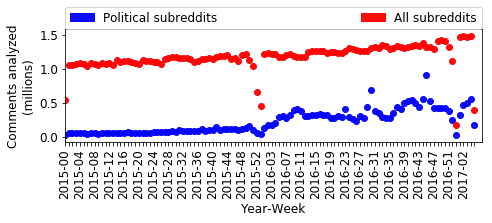}
\caption{Number of analyzed political and apolitical comments belonging to each
week between January 2015 and January 2017.}
\label{fig:comment-timeline}
\end{figure}

\myparab{Reddit comments dataset.} Finally, after building our offensive speech
classifier using the above datasets, we use it to classify comments made on
Reddit. While the complete Reddit dataset contains 2B comments made between the
period of January 2015 and January 2017, we only analyze only 168M. We select
comments to be analyzed using the following process: (1) we exclude comments
shorter than 10 characters in length, (2) we exclude comments made by
\texttt{[deleted]} authors, and (3) we randomly sample and include 10\% of all
remaining comments. We categorize comments made in any of 21 popular political
subreddits as \emph{political} and the remainder as \emph{apolitical}. Our final
dataset contains 129M apolitical and 39M political comments.
\Cref{fig:comment-timeline} shows the number of comments in our dataset that
were made during each week included in our study. We see an increasing number of
political comments per week starting in February 2016 -- the start of the 2016
US presidential primaries.

\section{Offensive Speech Classification}\label{sec:classifier}
In order to identify offensive speech, we propose a fully automated technique
that classifies comments into two classes: Offensive and Not Offensive. 

\subsection{Classification approach}
At a high-level, our approach works as follows:

\begin{itemize}
\item \textbf{Build a word embedding.} We construct a 100-dimensional word
embedding using all comments from our complete Reddit dataset (2B comments). 
\item \textbf{Construct a \emph{hate vector}.} We construct a list of offensive
and hateful words identified from external data and map them into a single
vector within the high-dimensional word embedding.
\item \textbf{Text transformation and classification.} Finally, we transform
text to be classified into scalars representing their distance from the
constructed hate vector and use these as input to a Random Forest classifier.
\end{itemize}

\myparab{Building a word embedding.} At a high-level, a word embedding maps
words into a high-dimensional continuous vector space in such a way that
semantic similarities between words are maintained. This mapping is achieved by
exploiting the distributional properties of words and their occurrences in the
input corpus.

Rather than using an off-the-shelf word embedding (\eg the GloVe embeddings
\cite{Pennington-2014} trained using public domain data sources such as
Wikipedia and news articles), we construct a 100-dimensional embedding using the
complete Reddit dataset (2B comments) as the input corpus. The constructed
embedding consists of over 400M unique words (words occurring less than 25 times
in the entire corpus are excluded) using the Word2Vec \cite{Mikolov-2013}
implementation provided by the Gensim library \cite{Rehurek-2010}. Prior to
constructing the embedding, we perform stop-word removal and lemmatize each word
in the input corpus using the SpaCy NLP framework \cite{Honnibal-2015}. The main
reason for building a custom embedding is to ensure that our embeddings capture
semantics specific to the data being measured (Reddit) -- \eg while the word
``\emph{karma}'' in the non-Reddit context may be associated with spirituality,
it is associated with points (comment and submission scores) on Reddit.

\myparab{Constructing a hate vector.} We use two lists of words associated with
hate \cite{Hatebase-API} and offensive \cite{Google-WDYL-data} speech to
construct a hate vector in our word embedding. This is done by mapping each word
in the list into the 100-dimensional embedding and computing the mean
vector. This vector represents the average of all known offensive words. The
main idea behind creating a hate vector is to capture the point (in our
embedding) to which the most offensive observed comments are likely to be near.
Although clustering our offensive word lists into similar groups and
constructing multiple hate vectors -- one for each cluster -- results in
marginally better accuracy for our classifier, we use this approach due to the
fact that our classification cost grows linearly with the number of hate vectors
-- \ie we need to perform $O(|S|)$ distance computations per hate vector to
classify string $S$. 

\begin{figure}[t]
\centering
\includegraphics[trim=0cm 0cm 0cm 0cm, clip=true, width=.49\textwidth, angle=0]
{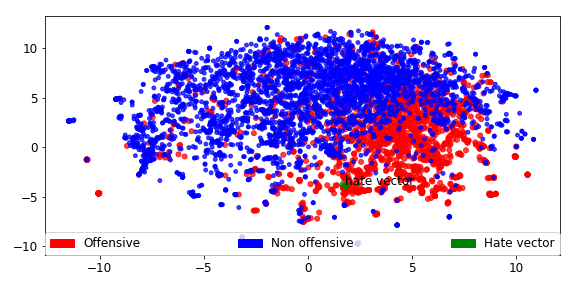}
\caption{Proximity of offensive and non-offensive comments to the hate vector.
Dimension reduction is performed using t-SNE.}
\label{fig:reduced-dimension-classes}
\end{figure}

\myparab{Transforming and classifying text.} We first remove stop-words and
perform lemmatization of each word in the text to be classified. We then obtain
the vector representing each word in the text and compute its similarity to the
constructed hate vector using the cosine similarity metric. A 0-vector is used
to represent words in the text that are not present in the embedding. Finally,
the maximum cosine similarity score is used to represent the comment.
Equation \ref{eq:text-xform} shows the transformation function on a string $S$ =
$\{s_1, \dots, s_n\}$ where $s_i$ is the vector representing the $i^{th}$
lemmatized non-stop-word, $\cos$ is the cosine-similarity function, and $H$ is
the hate vector. 
\begin{equation} 
T(S) = \max_{1 \leq i \leq n}[\cos(s_i, H)]
\label{eq:text-xform}
\end{equation}
In words, the numerical value assigned to a text is the cosine similarity
between the hate vector and the vector representing the word (in the text)
closest to the hate vector. This approach allows us to transform a string of
text into a single numerical value that captures its semantic similarity to the
most offensive comment. We use these scalars as input to a random forest
classifier to perform classification into Offensive and Not Offensive classes.
\Cref{fig:reduced-dimension-classes} shows the proximity of Offensive and Non
Offensive comments to our constructed hate vector after using t-distributed
Stochastic Neighbor Embedding (t-SNE) \cite{Maaten-2008} to reduce our
100-dimension vector space into 2 dimensions.

\subsection{Classifier evaluation}
We now present results to (1) validate our choice of classifier and (2)
demonstrate the impact of training/validation sample count on our classifiers
performance.

\begin{table}[h]
\centering
\resizebox{\linewidth}{!}{%
\begin{tab}{lll}
\textbf{Classifier} & \textbf{Accuracy (\%)} & \textbf{F1-Score (\%)} \\
\midrule
Stochastic Gradient Descent & 80.7 & 80.0 \\
Naive Bayes & 81.5 & 81.2 \\
Decision Tree & 91.8 & 91.4 \\
\textbf{Random Forest} & \textbf{92.0} & \textbf{91.9} \\
\end{tab}
}
\caption{Average classifier performance during 10-fold cross-validation on
the training/validation set. Results shown are for the best performing
parameters obtained using a grid search.}
\label{tab:classifiers}
\end{table}

\myparab{Classifier selection methodology.}
To identify the most suitable classifier for classifying the scalars associated
with each text, we perform evaluations using the stochastic gradient descent,
naive bayes, decision tree, and random forest classifiers. For each classifier,
we split the CrowdFlower hate speech dataset into a training/validation set
(75\%), and a holdout set (25\%). We perform 10-fold cross-validation on the
training/validation set to identify the best classifier model and parameters
(using a grid search). Based on the results of this evaluation, we select a
100-estimator entropy-based splitting random forest model as our classifier.
\Cref{tab:classifiers} shows the mean accuracy and F1-score for each evaluated
classifier during the 10-fold cross-validation. 

\myparab{Real-world classifier performance.}
To evaluate real-world performance of our selected classifier (\ie performance
in the absence of model and parameter bias), we perform classification of the
holdout set. On this set, our classifier had an accuracy and F1-score of 89.6\%
and 89.2\%, respectively. These results show that in addition to superior
accuracy during training and validation, our chosen classifier is also robust
against over-fitting.

\begin{figure}[tbp]
\centering
\begin{minipage}{.49\textwidth}
\begin{subfigure}{\textwidth}
\includegraphics[trim=0cm .5cm 0cm 0cm, clip=true, width=\textwidth, angle=0]
{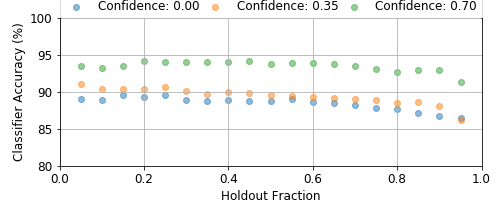}
\caption{Classifier accuracy.}
\label{fig:classifier-accuracy}
\end{subfigure}

\begin{subfigure}{\textwidth}
\includegraphics[trim=0cm .5cm 0cm 0cm, clip=true, width=\textwidth, angle=0]
{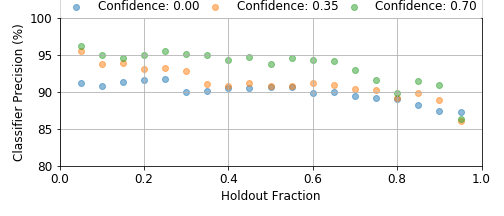}
\caption{Classifier precision.}
\label{fig:classifier-precision}
\end{subfigure}

\begin{subfigure}{\textwidth}
\includegraphics[trim=0cm .5cm 0cm 0cm, clip=true, width=\textwidth, angle=0]
{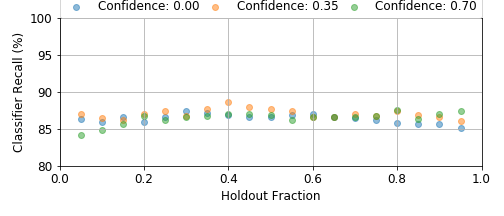}
\caption{Classifier recall.}
\label{fig:classifier-recall}
\end{subfigure}

\end{minipage}
\caption{Classifier performance on holdout sets while varying holdout set sizes
and minimum confidence thresholds.}
\label{fig:classifier-performance}
\end{figure}

\myparab{Impact of dataset quality and size.}
To understand how the performance of our chosen classifier model and parameters
are impacted by: (1) the quality and consistency of manually assigned classes in
the CrowdFlower dataset and (2) the size of the dataset, we re-evaluate the
classifier while only considering tweets having a minimum confidence score and
varying the size of the holdout set. Specifically, our experiments considered
confidence thresholds of 0 (all tweets considered), .35 (only tweets where at
least 35\% of contributors agreed on a class were considered), and .70 (only
tweets where  at least 70\% of the contributors agreed on a class were
considered) and varied the holdout set sizes between 5\% and 95\% of all tweets
meeting the confidence threshold set for the experiment.

The results illustrated in \Cref{fig:classifier-performance} show the
performance of the classifier while evaluating the corresponding holdout set. We
make several conclusions from these results:

\begin{itemize}
\item Beyond a (fairly low) threshold, the size of the training and validation
set has little impact on classifier performance. We see that the accuracy,
precision, and recall have, at best, marginal improvements with holdout set sizes
smaller than 60\%. This implies that the CrowdFlower dataset is sufficient for
building an offensive speech classifier.
\item Quality of manual labeling has a significant impact on the accuracy and
precision of the classifier. Using only tweets which had at least 70\% of
contributors agreeing on a class resulted in between 5-7\% higher accuracy and
up to 5\% higher precision.
\item Our classifier achieves precision of over 95\% and recall of over 85\%
when considering only high confidence samples. This implies that the classifier
is more likely to underestimate the presence of offensive speech -- \ie our
results likely provide a lower-bound on the quantity of observed offensive
speech.
\end{itemize}

\section{Measurements}\label{sec:measurement}
In this section we quantify and characterize offensiveness in the political and
general contexts using our offensive speech classifier and the Reddit comments
dataset which considers a random sample of comments made between January 2015
and January 2017.

\begin{figure}[ht]
\centering
\includegraphics[trim=0cm 0cm 0cm 0cm, clip=true, width=.49\textwidth, angle=0]
{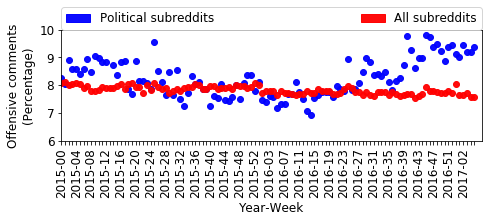}
\caption{Fraction of offensive comments identified in political and all
subreddits.}
\label{fig:offensive-speech-timeline}
\end{figure}

\myparab{Offensiveness over time.}
We find that on average 8.4\% of all political comments are offensive compared
to 7.8\% of all apolitical comments. \Cref{fig:offensive-speech-timeline}
illustrates the fraction of offensive political and apolitical comments made
during each week in our study. We see that while the fraction of apolitical
offensive comments has stayed steady, there has been an increase in the fraction
of offensive political comments starting in July 2016. Notably, this increase is
observed after the conclusion of the US presidential primaries and during the
period of the Democratic and Republican National Conventions and does not reduce
even after the conclusion of the US presidential elections held on November 8.
Participants in political subreddits were 2.6\% more likely to observe offensive
comments prior to July 2016 but 14.9\% more likely to observe offensive comments
from July 2016 onwards.

\begin{figure}[ht]
\centering
\includegraphics[trim=0cm 0cm 0cm 0cm, clip=true, width=.49\textwidth, angle=0]
{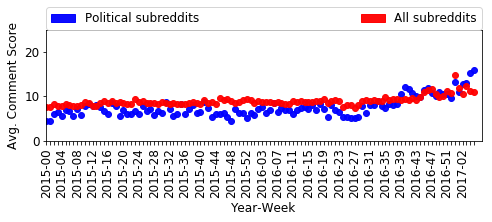}
\caption{Average scores of offensive comments identified in political and all
subreddits.}
\label{fig:offensive-scores-timeline}
\end{figure}

\myparab{Reactions to offensive comments.} We use the comment \emph{score},
roughly the difference between up-votes and down-votes received, as a proxy for
understanding how users reacted to offensive comments. We find that comments
that were offensive: (1) on average, had a higher score than non-offensive
comments (average scores: 8.9 \vs 6.7) and (2) were better received when they
were posted in the general context than in the political context (average
scores: 8.6 \vs 9.0). To understand how peoples reactions to offensive comments
evolved over time, \Cref{fig:offensive-scores-timeline} shows the average scores
received by offensive comments over time. Again, we observe an increasing trend
in average scores received by offensive and political comments after July 2016.

\begin{figure}[ht]
\centering
\includegraphics[trim=0cm 0cm 0cm 0cm, clip=true, width=.49\textwidth, angle=0]
{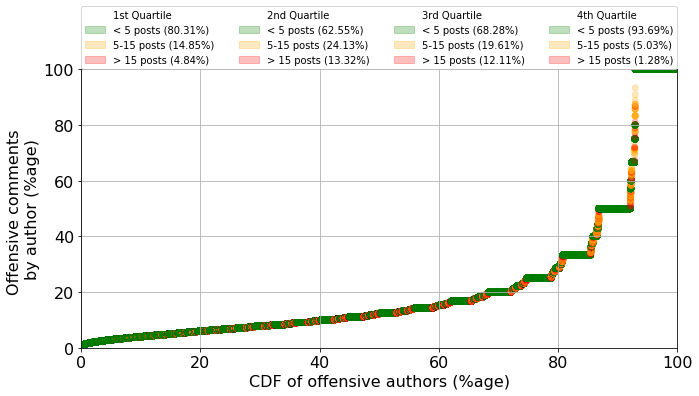}
\caption{CDF of the fraction of each authors comments that were identified as
offensive. Green, orange, and red dots are used to represent authors with $<$5,
5-15, and $>$15 total comments, respectively. The legend provides a breakdown
per quartile.}
\label{fig:offensive-authors-cdf}
\end{figure}

\myparab{Characteristics of offensive authors.}
We now focus on understanding characteristics of authors of offensive comments.
Specifically, we are interested in identifying the use of \emph{throwaway} and
\emph{troll} accounts. For the purposes of this study, we characterize
\emph{throwaway} accounts as those with less than five total comments -- \ie
accounts that are used to make a small number of comments. Similarly, we define 
\emph{troll} accounts as those with over 15 comments of which over 75\% are
classified as offensive -- \ie accounts that are used to make a larger number of
comments, of which a significant majority are offensive. We find that 93.7\% of
the accounts which have over 75\% of their comments tagged as offensive are
throwaways and 1.3\% are trolls. Complete results are illustrated in
\Cref{fig:offensive-authors-cdf}.

\begin{figure}[ht]
\centering
\includegraphics[trim=0cm 0cm 0cm 0cm, clip=true, width=.49\textwidth, angle=0]
{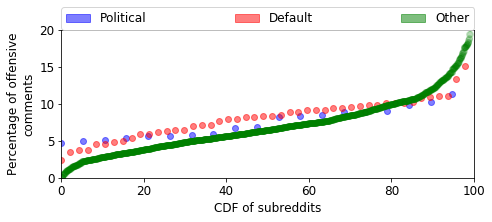}
\caption{Distribution of the fraction of offensive comments observed in each
subreddit category. Only subreddits with over 1000 comments are considered.}
\label{fig:subreddit-cdf}
\end{figure}

\myparab{Characteristics of offensive communities.} We breakdown subreddits
by their category (default, political, and other) and identify the most and
least offensive communities in each. \Cref{fig:subreddit-cdf} shows the
distribution of the fraction of offensive comments in each category and
\Cref{tab:subreddit-breakdown} shows the most and least offensive subreddits in
the political and default categories (we exclude the ``other'' category due to
the inappropriateness of their names). We find that less than 19\% of all
subreddits (that account for over 23\% of all comments) have over 10\%
offensive comments. Further, several default and political subreddits fall in
this category, including {\tt r/the$\_$donald} -- the most offensive political
subreddit and the subreddit dedicated to the US President. 

\begin{table}[th]
\centering
\resizebox{\linewidth}{!}{%
\begin{tab}{lll}
\textbf{Category} & \textbf{Most offensive (\%)} & \textbf{Least offensive (\%)} \\
\midrule
Default   & {\tt r/tifu} (15.1\%) 	   & {\tt r/askscience} (2.4\%) \\
	  & {\tt r/announcements} (13.2\%) & {\tt r/personalfinance} (3.4\%)\\
	  & {\tt r/askreddit} (11.0\%)	   & {\tt r/science} (3.8\%)\\\midrule
Political & {\tt r/the$\_$donald} (11.4\%) & {\tt r/republican} (4.4\%)\\
	  & {\tt r/elections} (10.2\%)	   & {\tt r/sandersforpresident} (4.9\%)\\
	  & {\tt r/worldpolitics} (9.8\%)  & {\tt r/tedcruz} (5.1\%)\\
\end{tab}
}
\caption{Subreddits in the default and political categories with the highest and
lowest fraction of offensive comments.}
\label{tab:subreddit-breakdown}
\end{table}

\begin{figure}[ht]
\centering
\includegraphics[trim=.5cm 2cm 0cm 0cm, clip=true, width=.5\textwidth, angle=0]
{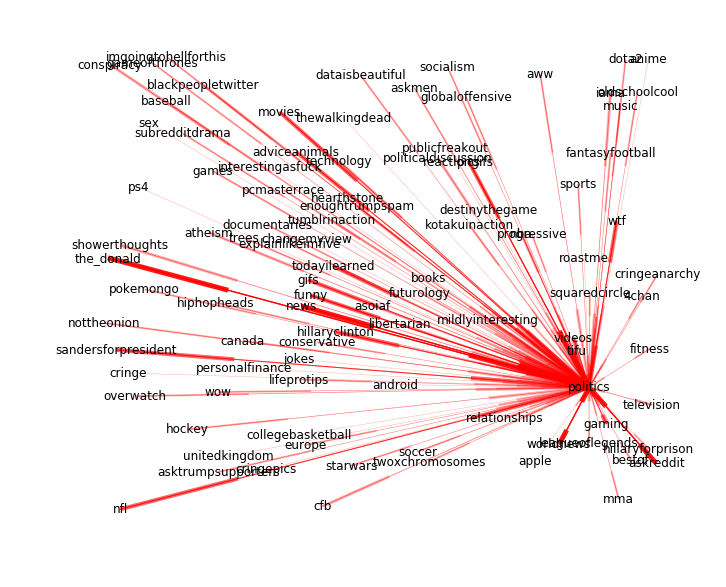}
\caption{Flow of offensive authors. An edge between two subreddits indicates
that authors made offensive comments in the source subreddit before the first
time they made offensive comments in the destination subreddit. Darker and
thicker edges indicate larger flow sizes (only flows $\ge 200$ authors are
shown).}
\label{fig:offensive-flow}
\end{figure}

\myparab{Flow of offensive authors.} Finally, we uncover patterns in the
movement of offensive authors between communities. In \Cref{fig:offensive-flow}
we show the communities in which large number of authors of offensive content
on the \texttt{r/politics} subreddit had previously made offensive comments (we
refer to these communities as sources). Unsurprisingly, the most popular sources
belonged to the default subreddits (\eg ~{\tt r/worldnews}, {\tt r/wtf}, {\tt
r/videos}, {\tt r/askreddit}, and {\tt r/news}). We find that several other
political subreddits also serve as large sources of offensive authors. In fact,
the subreddits dedicated to the three most popular US presidential candidates
-- {\tt r/the$\_$donald}, {\tt r/sandersforpresident}, and {\tt
r/hillaryclinton} rank in the top three. Finally, outside of the default and
political subreddits, we find that {\tt r/nfl}, {\tt r/conspiracy}, {\tt
r/dota2}, {\tt r/reactiongifs}, {\tt r/blackpeopletwitter}, and {\tt
r/imgoingtohellforthis} were the largest sources of offensive political authors.

\section{Conclusions and Future Work}\label{sec:conclusions}

We develop and validate an offensive speech classifier to quantify the presence
of offensive online comments from January 2015 through January 2017. We find
that political discussions on Reddit became increasingly less civil -- as
measured by the incidence of offensive comments -- during the 2016 general
election campaign. In fact, during the height of the campaign, nearly one of
every 10 comments posted on a political subreddit were classified as offensive.
Offensive comments also received more positive feedback from the community, even
though most of the accounts responsible for such comments appear to be
throwaway accounts. While offensive posts were increasingly common on political
subreddits as the campaign wore on, there was no such increase in non-political
fora. This contrast provides additional evidence that the increasing use of
offensive speech was directly related to the ramping up of the general election
campaign for president. 

Even though our study relies on just a single source of
online political discussions -- Reddit, we believe that our findings generally
present an upper-bound on the incidence of offensiveness in online political
discussions for the following reasons: First, Reddit allows the use of 
pseudonymous identities that enables the online disinhibition effect (unlike
social-media platforms such as Facebook). Second, Reddit enables users to engage
in complex discussions that are unrestricted in length (unlike Twitter).
Finally, Reddit is known for enabling a general culture of free speech and
delegating content regulation to moderators of individual subreddits. This
provides users holding fringe views a variety of subreddits in which their
content is welcome.

Our findings provide a unique and important mapping of the increasing incivility
of online political discourse during the 2016 campaign. Such an investigation is
important because scholars have outlined many consequences for incivility in
political discourse. Incivility tends to ``turn off'' political moderates,
leading to increasing polarization among those who are actively engaged in
politics \cite{wolf2012incivility}. More importantly, a lack of civility in
political discussions generally reduces the degree to which people view
opponents as holding legitimate viewpoints. This dynamic makes it difficult for
people to find common ground with those who disagree with them
\cite{mutz2006hearing} and it may ultimately lead citizens to view electoral
victories by opponents as lacking legitimacy \cite{mutz2015your}. Thus, from a
normative standpoint, the fact that the 2016 campaign sparked a marked increase
in the offensiveness of political comments posted to Reddit is of concern in its
own right; that the incidence of offensive political comments has remained high
even three months after the election is all the more troubling.  

In future work, we will extend our analysis of Reddit back to 2007 with the aim
of formulating a more complete understanding of the dynamics of political
incivility. For example, we seek to understand whether the high incidence of
offensive speech we find in 2016 is unique to this particular campaign or if
previous presidential campaigns witnessed similar spikes in incivility. We will
also examine whether there is a more general long-term trend toward offensive
online political speech, which would be consistent with what scholars have found
when studying political elites \cite{shea2012rise, jamieson2000continuity}.

\bibliographystyle{plain}
\bibliography{bibliography}

\end{document}